\documentclass{article}

\usepackage[preprint]{neurips_2025}

\usepackage[utf8]{inputenc} 
\usepackage[T1]{fontenc}    
\usepackage{hyperref}       
\usepackage{url}            
\usepackage{booktabs}       
\usepackage{amsfonts}       
\usepackage{nicefrac}       
\usepackage{microtype}      
\usepackage{xcolor}         
\usepackage{graphicx}
\usepackage{bm, amsmath}
\usepackage{ amssymb }
\usepackage[noabbrev, capitalize]{cleveref}

\title{Brain2Model Transfer: Training sensory and decision models with human neural activity as a teacher}

\author{%
  Tomas Gallo Aquino\\
  Columbia University\\
  New York, NY 10027\\
  \texttt{tg2863@columbia.edu} \\
  \And
  Victoria Liu\\
  Columbia University\\
  New York, NY 10027\\
  \texttt{ql2491@columbia.edu} \\
  \And
  Habiba Azab\\
  Baylor College of Medicine\\
  Houston, TX 77030\\
  \texttt{azab@bcm.edu} \\
  \And
  Raissa Mathura\\
  Baylor College of Medicine\\
  Houston, TX 77030\\
  \texttt{Raissa.Mathura@bcm.edu} \\
  \And
  Andrew J. Watrous\\
  Baylor College of Medicine\\
  Houston, TX 77030\\
  \texttt{andrew.watrous@bcm.edu} \\
  \And
  Eleonora Bartoli\\
  Baylor College of Medicine\\
  Houston, TX 77030\\
  \texttt{bartoli@bcm.edu} \\  
  \And
  Benjamin Y. Hayden\\
  Baylor College of Medicine\\
  Houston, TX 77030\\
  \texttt{benjamin.Hayden@bcm.edu} \\
  \And
  Paul Sajda\\
  Columbia University\\
  New York, NY 10027\\
  \texttt{ps629@columbia.edu} \\
  \And
  Sameer A. Sheth*\\
  Baylor College of Medicine\\
  Houston, TX 77030\\
  \texttt{sameer.Sheth@bcm.edu} \\
\And
  Nuttida Rungratsameetaweemana*\\
  Columbia University\\
  New York, NY 10027\\
  \texttt{nr2869@columbia.edu} \\
}

\begin{document}

\maketitle

\begin{abstract}
Transfer learning enhances the training of novel sensory and decision models by employing rich feature representations from large, pre-trained teacher models. Cognitive neuroscience shows that the human brain creates low-dimensional, abstract representations for efficient sensorimotor coding. Importantly, the brain can learn these representations with significantly fewer data points and less computational power than artificial models require. We introduce Brain2Model Transfer Learning (B2M), a framework where neural activity from human sensory and decision-making tasks acts as the teacher model for training artificial neural networks. We propose two B2M strategies: (1) Brain Contrastive Transfer, which aligns brain activity and network activations through a contrastive objective; and (2) Brain Latent Transfer, which projects latent dynamics from similar cognitive tasks onto student networks via supervised regression of brain-derived features. We validate B2M in memory-based decision-making with a recurrent neural network and scene reconstruction for autonomous driving with a variational autoencoder. The results show that student networks benefiting from brain-based transfer converge faster and achieve higher predictive accuracy than networks trained in isolation. Our findings indicate that the brain's representations are valuable for artificial learners, paving the way for more efficient learning of complex decision-making representations, which would be costly or slow through purely artificial training.

\small
\vspace{15mm}
$^{*}$ These authors jointly supervised this work.
\normalsize
\end{abstract}


\section{Introduction}

Recent work in computational neuroscience reveals that neural populations encode low-dimensional, abstract representations that enhance decision-making and sensorimotor behavior \cite{sussillo2015neural,gao2015simplicity,gallego2017neural,bernardi2020geometry, johnston2024semi}. A key objective in the field is to create models that predict neural activity while uncovering the computational principles behind these representations. Advances have been achieved in vision and language, where convolutional and transformer-based models elucidate neural responses in corresponding brain areas \cite{yamins2014performance,tang2023brain,tang2023semantic,margalit2024unifying,cao2024explanatory}. These findings indicate that artificial neural networks and biological systems may adopt similar representational strategies for perception and decision-making.

Despite advances in large-scale artificial models in vision \cite{he2016deep,howard2019searching,radford2021learning,oquab2023dinov2}, language \cite{reimers2019sentence,barbieri2020tweeteval,achiam2023gpt,touvron2023llama,liu2024deepseek}, and decision-making \cite{mnih2015human,silver2017mastering,silver2018general}, training neural networks to achieve generalizable abstractions typically demands extensive supervision or large foundation models, which are slow and costly to obtain. In contrast, the brain learns complex representations from sparse and ambiguous data, efficiently guiding flexible decision-making and enabling rapid adaptation in novel settings \cite{lake2017building,lake2019human}. 

Cognitive neuroscience has revealed how the brain creates low-dimensional task-relevant representations. However, most previous methods have not effectively used these representations to influence the learning dynamics of artificial models.

\section{Related Work}
Leveraging representations from large pre-trained models for efficient training of new task-specific models has been extensively explored in transfer learning literature, achieving success in both supervised \cite{wang2011heterogeneous,duan2012learning,zhou2014hybrid,weiss2016survey,tan2018survey} and reinforcement learning \cite{taylor2007transfer,kim2013learning,czarnecki2019distilling,zhu2023transfer}. 

Different transfer approaches have been proposed based on the type of knowledge being transferred and the differences between the data available to the source and target models \cite{5288526,weiss2016survey,tan2018survey}, facilitating knowledge transfer across complex decision domains. 

Past work has explored using sensory-based brain information to train convolutional neural networks (CNNs) \cite{mcclure2016representational,fong2018using,nishida2020brain}. In McClure and Kriegeskorte (2016, the authors defined representational dissimilarity matrices (DSMs) leveraging data in an MNIST/CIFAR100 visual recognition task and utilized them to transfer information from a teacher model, which could in theory be a brain, to a student CNN. Fong et. al, 2018 have shown the possibility of guiding support vector machine image classifiers with brain-derived features. In Nishida et al. (2020, the authors utilize a series of linear regression models to predict voxel-wise functional magnetic resonance imaging (fMRI) responses during audiovisual stimulus presentation and subsequently guide learning of a VGG-16 model in predicting scene-derived labels. 

Human-derived behavioral variables may also be leveraged in this fashion to guide neural network training \cite{shih2017towards,sajda2023systems,sucholutsky2023alignment}. For instance, in Sajda et al. (2023), the authors propose utilizing behavioral variables such as level of interest, arousal, emotional reactivity, cognitive fatigue, or cognitive state for training deep reinforcement learning models. Similarly, Shih et al. (2017) discuss the interaction of these factors with electroencephalography (EEG) recordings in adapting AI systems to human idiosyncrasies. Additionally, in Sucholutsky and Griffiths (2023), the authors show how even aligning vision models to human behavioral judgments can improve their generalization, few-shot learning, and robustness to adversarial attacks.

Despite this relevant work in the field, some outstanding gaps remain regarding the flexibility of these methods and their application domains. For instance, utilizing representational dissimilarity matrices to approximate neural data and artificial models requires discrete DSM classes to be manually defined a priori, which reduces the flexibility of this method, especially for applications with non-discrete state/action spaces. Additionally, leveraging higher-order cognitive functions as transfer signals, such as memory or decision making, remains an underexplored research direction, above and beyond convolutional neural networks for vision. Developing a general brain transfer learning framework aiming to maximize mutual information between neural and artificial systems would be an important step in this direction.


\section{Brain2Model Transfer Learning methods}
To address these gaps, we propose Brain2Model Transfer Learning (B2M) to improve artificial model learning by encouraging models to achieve similar embedding representations to the human brain, leveraging human brain data collected while subjects performed the same task learned by the model. Concretely, we augment standard learning objectives by adding a brain transfer loss term:

\begin{equation}
    \mathcal{L}_{total} = (1-\alpha) \mathcal{L}_{task} + \alpha \mathcal{L}_{transfer}
\end{equation}

where $\mathcal{L}_{task}$ is the standard loss function for the artificial learner in the task, $\mathcal{L}_{transfer}$ is the loss function comparing brain and artificial representational similarity, and $\alpha$ is the transfer weight hyperparameter, designed to trade-off between transfer loss and task loss. 

\begin{figure}[htb!]
  \centering
  \includegraphics[width=1\textwidth]{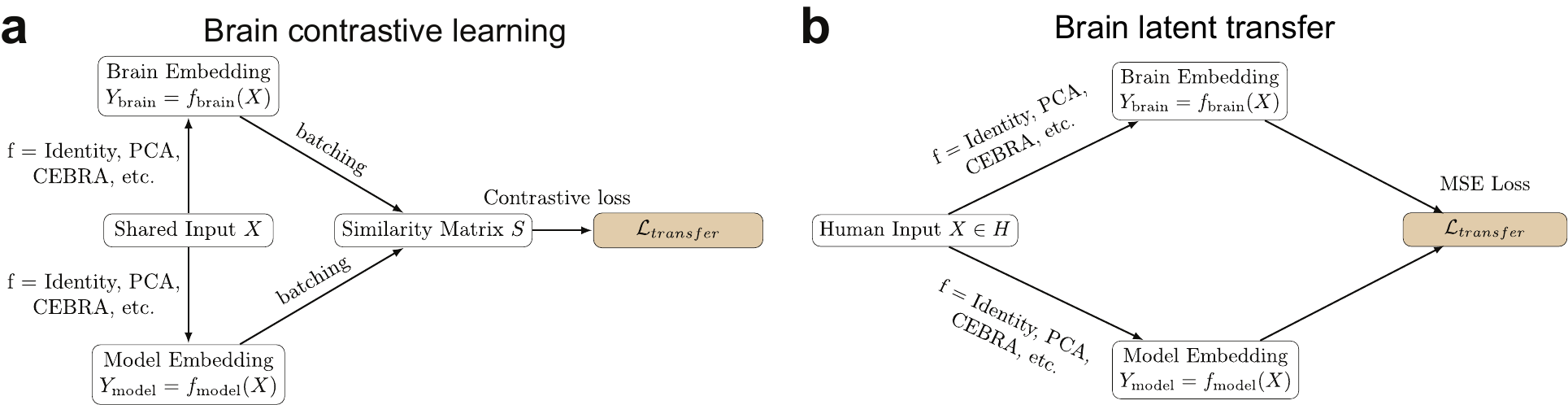}
 \caption{Obtaining $\mathcal{L}_{transfer}$ in B2M. (a) Brain contrastive learning. A shared input is transformed into both brain embeddings and artificial model embeddings of equal dimensionality. From input batches containing several examples, we compute a similarity matrix between brain and model embeddings of different inputs, which is utilized to obtain a transfer loss value. We designate neural-model embedding pairs computed from the same input example as positive pairs, while pairs computed from separate examples are designated negative pairs. (b) Brain latent transfer learning. An input $X$ from the human train set $\mathcal{H}$ is used to produce brain embeddings and artificial embeddings, produced from a model trained with an artificial train set $\mathcal{A}$. These embeddings are subsequently compared with a mean square error distance to obtain a transfer loss.}
  \label{fig:diagram}
\end{figure}

Since this is a general framework, obtaining $\mathcal{L}_{transfer}$ is dependent on the current task and data collection setup, and on the extent to which brain data and model training sets are aligned. Therefore, we propose the following two alignment strategies, whose applicability depends on whether artificial models are trained exactly on the same examples as seen by humans who performed the task, or approximations of these examples (Fig. \ref{fig:diagram}).

\subsection{Brain Contrastive Loss}
One goal of B2M is to maximize the mutual information between neural sensorimotor representations and their counterparts in artificial models. This aims to assist artificial learning by encouraging models to find advantageous sensorimotor representational subspaces that might have been acquired by the human brain over the course of task learning and, ultimately, evolution. We propose to achieve this by adapting contrastive learning via an InfoNCE loss framework, such as the one adopted in SimCLR, \cite{oord2018representation, chen2020simple}, which aims to maximize mutual information between similar data points (Fig. \ref{fig:diagram}a). 

Given a sensory or contextual input $X \in \mathcal{I}$, with $X \in \mathbb{R}^{d_{1}\times d_{2} \times ...\times d_{n}}$, where $\mathcal{I}$ is a train set presented to both humans and artificial models, we propose aligning brain and artificial representations of X by maximizing the mutual information between embeddings $Y_{brain} \in \mathbb{R}^{E}$ and $Y_{model} \in \mathbb{R}^{E}$, where $E$ is the embedding dimensionality. These embeddings are produced by non-linear transfer functions such that $Y_{brain} = f_{brain}(X)$ and $Y_{model} = f_{model}(X)$, indicating compressed neural and artificial representations of sensorimotor inputs, respectively. Any applicable dimensionality reduction method can be utilized as $f$, or even the identity function, as long as the dimensionality between brain and model embeddings is matched.

For this, we define a batch of $b$ examples  $X \in \mathcal{I}$, $[X_{1},X_{2},...,X_{b}]$, and their respective neural and artificial embeddings $B_{neural} = [Y_{(neural,1)},Y_{(neural,2)},...,Y_{(neural,b)}]$ and $B_{model} = [Y_{(model,1)},Y_{(model,2)},...,Y_{(model,b)}]$. From these, given a temperature hyperparameter $\tau$, we obtain a similarity matrix $\mathcal{S} = \frac{1}{\tau}(B_{neural} \times B_{model}^{T})$, which represents the similarity between neural and artificial example pairs, including both pairs obtained from the same example $X_{i}$, but also pairs obtained from different examples. Then, we define positive contrastive pairs $Y_{(neural,i)}$ and $Y_{(model,i)}$, and negative contrastive pairs $Y_{(neural,i)}$ and $Y_{(model,j)}$ for all $i \neq j$. With these, we define, for a given anchor example $i$, with $\mathcal{L}_{transfer} = \sum_{i}\mathcal{L}_{transfer,i}$:

\begin{equation}
    \mathcal{L}_{transfer,i} = -\log \frac{exp(S_{i,i})}{exp(S_{i,i}) + \sum_{i \neq j} exp(S_{i,j})}
\end{equation}

\subsection{Brain Latent Transfer Loss}
In decision-making tasks, future states are often dependent on decisions made in past episodes, as well as the outcomes that occurred in them. For this reason, it is potentially challenging to assemble a train set of episodes and environment states for an artificial agent that will exactly match the train sets experienced by human subjects in a task, given that the agent is free to act differently from their human counterparts during training. To circumvent this, instead of exact episode matching, we propose matching brain embeddings obtained during exposure to an input $X$ to the artificial embeddings obtained during exposure to the same input $X$, as long as $X$ is an approximation of the examples contained in the artificial train set (Fig. \ref{fig:diagram}b). 

Concretely, given a sensory or contextual input $X \in \mathcal{H}$, with $X \in \mathbb{R}^{d_{1}\times d_{2} \times ...\times d_{n}}$, where $\mathcal{H}$ is the train set presented to humans, we assume brain activity produces an embedding $Y_{brain}$ via a non-linear transfer function $f_{brain}$, such that $Y_{brain} = f_{brain}(X)$, with $Y_{brain} \in \mathbb{R}^{E}$, in which $E$ is the embedding dimension. Additionally, we assume the artificial model learns from examples $X' \in \mathcal{A}$, $X' \in \mathbb{R}^{d_{1}\times d_{2} \times ...\times d_{n}}$, where $\mathcal{A}$ is the train set presented to the artificial model, related but not necessarily equal to $\mathcal{H}$. Throughout its learning process, the artificial model finds a non-linear transfer function $f_{model}$ which produces its own embedding $Y_{model}' = f_{model}(X')$, with $Y_{model}' \in \mathbb{R}^{E}$. 

Then, we propose to achieve brain-to-model transfer by performing a latent transfer between brain and model embeddings. Concretely, for an input $X \in \mathcal{H}$ previously presented to humans, we obtain its embedding produced by the artificial model $Y_{model} = f_{model}(X)$ and minimize its mean square error distance to its known corresponding brain embedding $Y_{brain} = f_{brain}(X)$:

\begin{equation}
    \mathcal{L}_{transfer} = \frac{1}{E} \sum^{E}_{i=1}(Y_{model,i} - Y_{brain,i})^{2}
\end{equation}

\section{B2M improves RNN learning in memory-based decision making task}

\subsection{Memory task and brain embeddings}

\begin{figure}[htb!]
  \centering
  \includegraphics[width=0.85\textwidth]{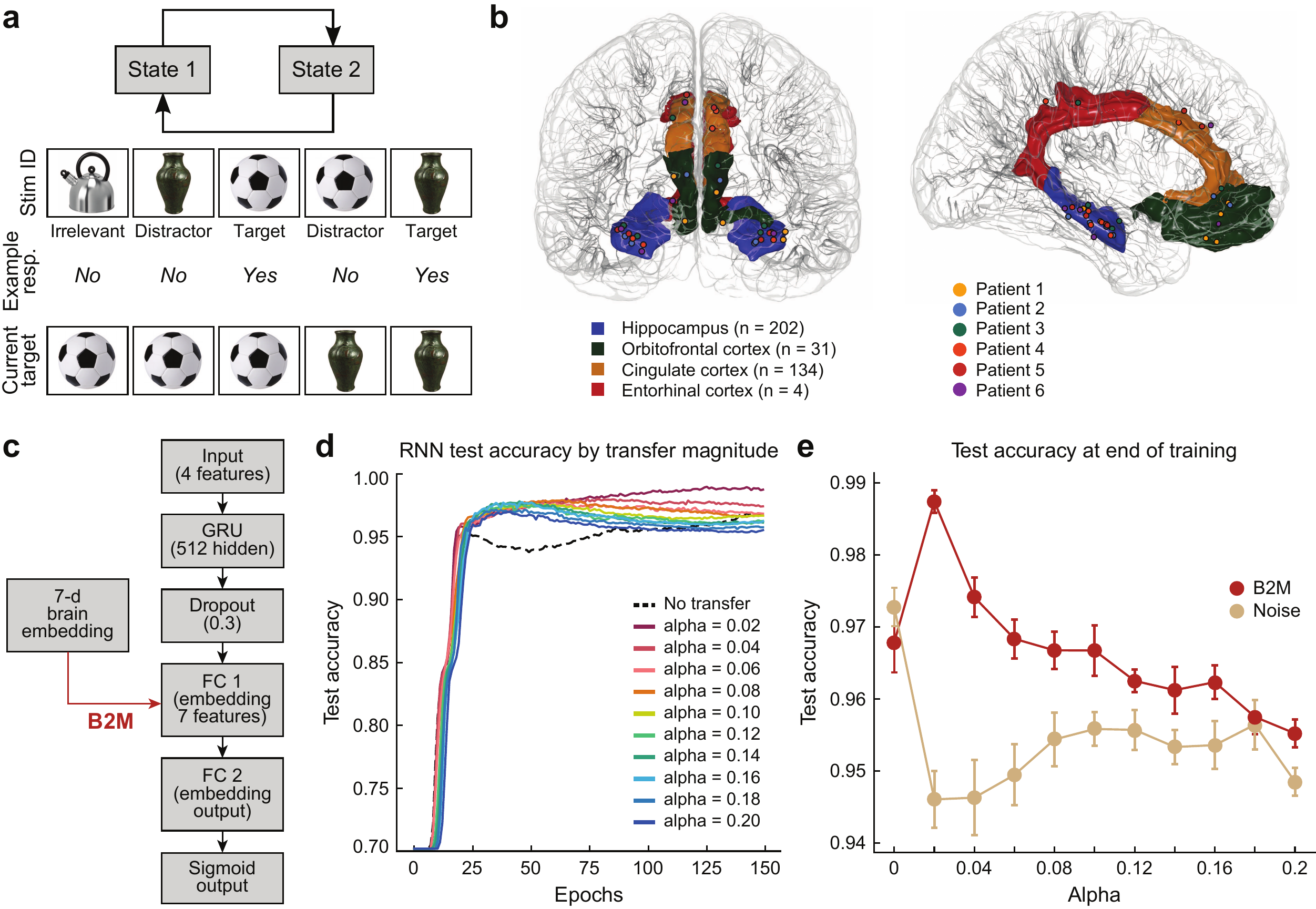}
 \caption{RNN performance in memory-based decision making task with B2M. (a) Task design. The task alternated between two states, indicating which of the stimuli was the current target. In each episode, two stimuli were designated as potential targets, and one stimulus was always irrelevant. Patients observed a sequence of stimulus images and were prompted to respond whether the current image was the target or not. The task alternated to the other state once a target was correctly identified. (b) Invasive electrode mapping for single neuron recording across epilepsy patients. (c) RNN for memory-based decision making. The agent consisted of a GRU layer with dropout, followed by two fully connected (FC) layers. The first FC layer was jointly embedded with human brains via B2M. (d) Mean RNN test accuracy over epochs for different values of B2M strength $\alpha$, including no B2M (dashed black curve). (e) Test accuracy at the end of training for different values of B2M strength $\alpha$, for true brain embeddings (red) and noise (beige). Error bars represent standard error of the mean. }
  \label{fig:rnn_results}
\end{figure} 

The flexibility of human cognition remains a challenge for artificial systems to fully replicate. Given novel contexts or a novel instruction set, humans are able to quickly adapt and correctly follow new goals \cite{smith2019widespread, henderson2025dynamic, katlowitz2025learning, aquino2023neurons}. To understand the role of human brain activity in rapid goal switching, we utilized data collected from a memory-based goal switching task (Fig. \ref{fig:rnn_results}a) in which intracranially implanted epilepsy patients (N=17 sessions in 6 patients) had to memorize the current target stimulus and watch a sequence of visual stimuli. If the current target was presented, the correct response would be to press an accept button; otherwise, the correct response would be to press a reject button. As soon as the subject correctly accepted the current target, the target stimulus would shift to another stimulus in the episode, and the current target would alternate between these two until the end of the episode. Episodes had a variable number of steps (stimulus presentations) and 3 possible stimuli: 2 target candidates and one distractor stimulus. In total, 400 steps were given to each subject. 

Invasive neural data was acquired with IRB approval and informed consent, utilizing the Blackrock system. Standard spike sorting was performed with WaveClus \cite{quiroga2004unsupervised}. Following spike sorting, we obtained a total of 371 neurons (31 in orbitofrontal cortex, 202 in hippocampus, 4 in entorhinal cortex, and 134 in cingulate cortex, Fig. \ref{fig:rnn_results}b). We applied the following pre-processing steps to temporally align neural data and RNN activity: (1) we aligned all spikes to stimulus presentation and created two time periods: pre-stimulus (-1s to 0, resolution: 10ms) and post-stimulus (0s to reaction time). (2) We normalized post-stimulus spikes by reaction time to always fit into a 100-element rate vector, and concatenated pre-stimulus and post-stimulus normalized spikes into a spiking rate vector. (3) We filtered spike rates by convolving the rate vector of each step with a causal exponential filter (kernel: 20 zeros followed by $e^{-0.5x}$, $x = [0,0.5,1,...,9.5,10]$).

With this neural dataset, our study focused on transfer learning between behavioral sequences and neural representations in this structured sequential decision-making task. Training inputs were comprised of one-hot coded stimulus identities, organized into temporally ordered features as task events, and corresponding binary target sequences.

To learn shared representations across multiple behavioral sessions, we used the Contrastive Embedding by Relative Arrangement (CEBRA) framework \cite{schneider2023learnable} (Apache License). CEBRA is a self-supervised learning algorithm that maps high-dimensional neural activity to a low-dimensional space by leveraging temporal structure and optional contextual supervision. Below, we describe the full procedure used to generate multi-session embeddings from neural recordings across 17 behavioral sessions. 

We aggregated neural activity from all neurons in the 17 distinct recording sessions. Each session contained neural population firing rates stored as 3D arrays with shape $(N_{steps},N_{times},N_{neurons})$, where $N_{steps}$ represents the number of unique stimuli presented to a subject, considering each stimulus constitutes a step. For each session, data were reshaped into 2D arrays of shape $(N_{steps} \times N_{times},N_{neurons})$ and paired with time labels repeated across trials. These reshaped arrays represent temporally ordered sequences of neural activity and serve as the input to CEBRA.

We instantiated a CEBRA model configured for multi-session contrastive learning. The following hyperparameters were used: model architecture: offset10-model, batch size: 512, learning rate: $3 \cdot 10^{-4}$, temperature mode: auto with a minimum temperature of 0.1, embedding dimensionality: 7, maximum training iterations: 15,000, distance metric: cosine similarity, supervision: conditional sampling based on relative time (i.e., "time-delta"), utilizing time within step as a supervising feature.

We trained the model jointly on all 17 sessions by supplying the data as a list of session-wise arrays. This procedure allows the model to learn an embedding space that generalizes across session boundaries while respecting the temporal dynamics within each session. After training, each session's neural data were projected into a shared 7-dimensional embedding space. This dimensionality was chosen since it was the minimum available number of neurons in any given session. The output consisted of one embedding per time point per trial for each session.

\subsection{Model architecture and training}
To model an artificial agent performing this task, we implemented a single-layer gated recurrent unit (GRU) network designed for sequence modeling with alignment to brain-derived embeddings (Fig. \ref{fig:rnn_results}c). Input size was 4, to provide a one-hot coded version of the 3 possible stimuli, plus one element for when no stimulus was on the screen (i.e., pre-stimulus window). The architecture consisted of a GRU layer (hidden size=512, num layers=1), a dropout layer (rate 0.3), transfer temperature = 0.1, a linear embedding projection layer (hidden dimensions: 7), an output layer mapping the embedding to a single continuous output, followed by a scaled sigmoid activation mapping outputs to the range [-1.5, 1.5]. This structure allowed the model to jointly produce behaviorally predictive outputs\textbf{ while internally encoding representations that could align with neural embeddings, via direct brain contrastive learning}. In training with brain alignment, we presented trial sequences to the network in a one-to-one mapping with episodes seen by human subjects, by one-hot coding the presented stimulus and showing the context of each episode to the model as a one-hot coded version of the current target stimulus at the beginning of each episode. We trained the network for the same total number of steps seen by humans (6800 steps in total across sessions). Subsequently, we tested network performance in 1000 simulated sequences of 26 steps each, totaling 26000 test steps. The length 26 was chosen since it was the maximum sequence length experienced by humans in the task. These sequences were previously unseen by humans. Simulated data was created with the same task rules as the real experiment.

Each configuration (i.e., each value of $\alpha = [0,0.02,...,0.2]$) was trained with 10 random initialization seeds. Models were optimized with Adam (learning rate = $10^{-4}$) for 150 epochs with a batch size of 1. Model performance was evaluated on held-out simulated test episode data. Accuracy was measured by comparing predictions to binary targets.

\subsection{Results}

We measured network performance by accuracy in left-out trials, determined by whether the agent correctly accepted or rejected stimuli in each step (Fig. \ref{fig:rnn_results}d). While all non-zero magnitudes of brain transfer $\alpha$ led to faster approximation to convergence values, we also observe that an optimal magnitude of $\alpha$ ($\alpha = 0.02$) led to significantly better performance at the end of training ($accuracy = 0.987$), than in the absence of brain transfer, at $accuracy = 0.967$ ($p<0.001$, Wilcoxon rank-sum test, one-sided), (Fig. \ref{fig:rnn_results}e). The displayed standard error was computed across 10 initialization seeds. To address the possibility that the network might be learning better with B2M than no-transfer due to noise regularization \cite{camuto2020explicit, rungratsameetaweemana2025random}, we performed 10 control runs for each value of $\alpha$, changing brain embeddings for standard Gaussian noise. We observe that noise-driven learning actually performs worse than no-transfer and, consequently, worse than B2M as well (Fig. \ref{fig:rnn_results}e).

\section{B2M for naturalistic scene reconstruction in VAE for driving task}

\subsection{Virtual reality driving task and EEG recordings}

\begin{figure}[htb!]
  \centering
  \includegraphics[width=1\textwidth]{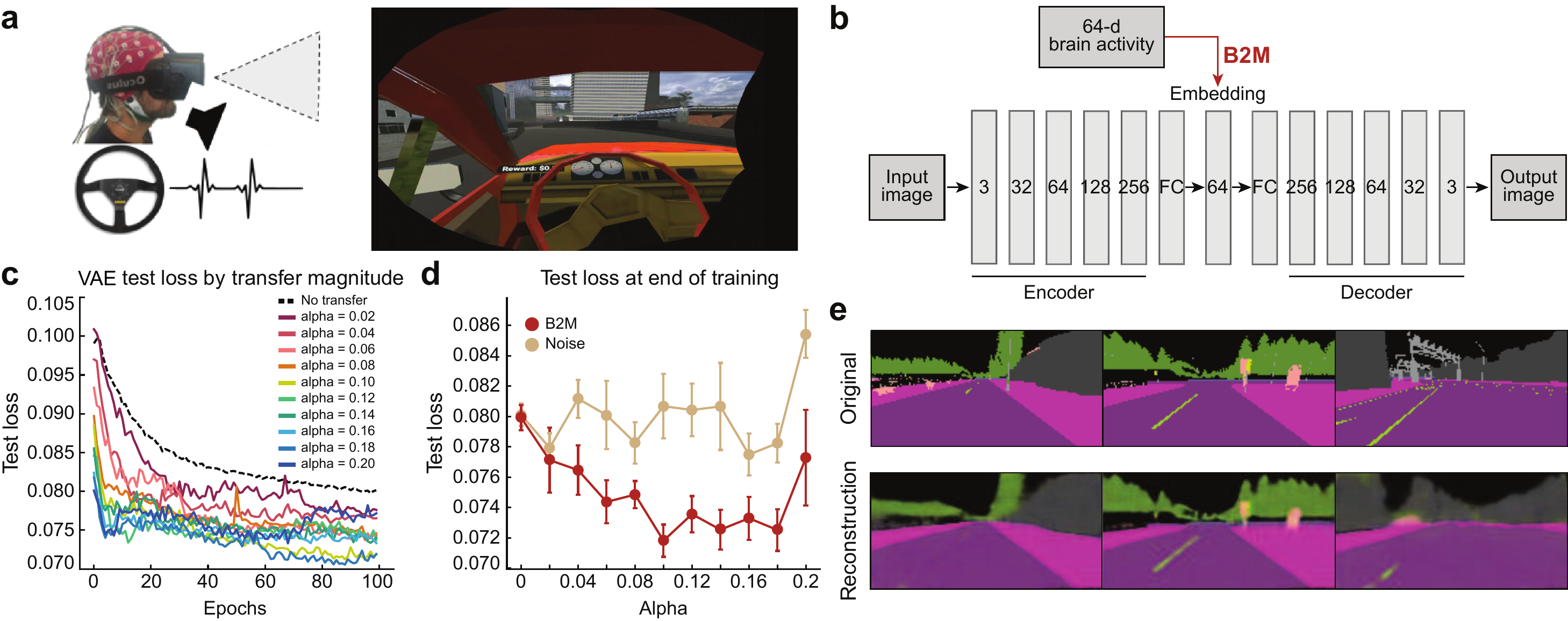}
 \caption{VAE performance in driving scene reconstruction with B2M. (a) Left: Data collection schematic: subjects steered a vehicle in VR with simultaneous EEG recordings; Right: VR driving scene example, as experienced by human subjects. (b) The VAE for scene reconstruction consisted of an encoder-embedding-decoder architecture. The embedding layer was jointly embedded with EEG data from human subjects driving in VR. (c) Mean test loss over epochs for different values of B2M strength $\alpha$, including no B2M (dashed black curve). (d) Test loss at the end of training for different values of B2M strength $\alpha$, for B2M with true brain data (red) and noise (beige). Error bars represent standard error of the mean. (e) Examples of original (top) and reconstructed (bottom) vehicle driving scenes, produced with B2M trained with $\alpha=0.1$. }
  \label{fig:driving_results}
\end{figure} 

To evaluate the generality and scalability of the B2M framework beyond invasive recordings, we next turn to a naturalistic visual reconstruction task using non-invasive EEG data. We explore B2M in a distinct model architecture (variational autoencoders) and apply the Brain Latent Transfer strategy to accommodate the lack of one-to-one trial correspondence between human and artificial data. This enables us to assess the effectiveness of B2M in a more ecologically valid and scalable setting involving real-time visual processing. For this, we tested B2M on a dataset of human subjects performing a vehicle driving task in a virtual reality (VR) environment, while they had 64-channel EEG activity recorded (Fig. \ref{fig:driving_results}a). This task has been previously shown to elicit activity in dorsolateral prefrontal cortex and anterior cingulate that are predictive of vehicle steering behavior \cite{koorathota2023pupil}. 

Healthy adults (N=11 sessions in 9 subjects, with written informed consent, IRB approved) completed a boundary‑avoidance driving task inside a virtual‑city environment rendered through an HTC Vive Pro Eye headset. Seated at a Logitech G920 steering wheel with accelerator and brake pedals, participants piloted a virtual car while continuous “fog” opacity dynamically modulated visual uncertainty on a trial‑by‑trial staircase. Crashes with road boundaries incurred point penalties to encourage timely, accurate steering. Scalp EEG was recorded throughout with a 64‑channel BioSemi ActiveTwo system (Ag/AgCl active electrodes, international 10–20 system, $impedances < 50 k\Omega$) at 2048Hz; a lossless screen‑capture of the VR scene was recorded via the Unity engine; steering‑wheel position, pedal inputs, and headset‑embedded eye‑tracking data were time‑synchronized with the EEG stream for later source and connectivity analyses. 

One possible task to model in this setting is the sensorimotor processing that occurs as the brain is parsing out a visual scene into actionable items, with the goal of safely driving through the city and accruing rewards. As such, we adapt a previously established \cite{razak2022implementing} variational autoencoder (VAE) trained with the specific purpose of creating an effective embedding of urban scenes for driving, to be later passed onto a reinforcement learning agent. The VAE (MIT License) has already been described elsewhere, but we will detail its adaptation here.

The model consists of an encoder-embedding-decoder architecture (Fig. \ref{fig:driving_results}b), that reconstructs input target urban scenes into matching outputs. These scenes were previously obtained in the CARLA driving simulator environment \cite{Dosovitskiy17} and made publicly available \cite{razak2022implementing}, with 12000 preset training images and 2000 test images, which are examples of driving scenes. We changed the embedding dimensionality from the original VAE to 64 dimensions, to directly match the 64 channels recorded in EEG sessions. The encoder and decoder each contain 5 convolutional/leaky ReLU layers and one fully connected layer (encoder dimensions: $[3,32,64,128,256]$, decoder dimensions: $[256,128,64,32,3]$, fully connected dimensions: $1024$.). The last convolutional encoder layer also contains a batch normalization step. Training visual scenes to be reconstructed were presented in batches (batch size: 32) of $160 \times 80$ pixel images.

For transfer learning, we used temporally aligned brain signals (EEG) and downsampled VR video frames (8Hz), reshaped to $160\times 80$ pixels, totaling 133120 frames across all subjects. \textbf{Since the VAE train set (simulated environment) and the human train set in VR are not exactly matching, we employed Brain Latent Transfer for B2M}. Additionally, since all subjects had the same spatially matching EEG channels, we did not perform any additional brain embedding steps. To present video/EEG data to the model in order to build the transfer loss, we treated each temporally aligned video/EEG example pair as a unique input and batched them (batch size: 256) and presented one batch at a time, in tandem with each training image batch.

All network hyperparameters were kept the same as in the original implementation \cite{razak2022implementing}, except for embedding dimensionality, which we changed to 64, in alignment with EEG inputs. Training was repeated across 10 random initializations for each $\alpha = [0,0.02,...,0.2]$ value, yielding a total of 110 full training runs. Each model was trained for 100 epochs, using the Adam optimizer (learning rate = $10^{-4}$) and default momentum settings. After each epoch, models were evaluated on a held-out validation set of naturalistic scenes.

\subsection{Results}
In a small percentage of runs, we observed diverging test loss ($loss > 1$) that did not recover over the course of training (5 out of 110 runs, $4.5\%$). We excluded these runs from the subsequent analysis and visualization. Further work must be done to characterize the instances in which diverging learning occurs.

We measured network performance by reconstruction loss in left-out naturalistic driving scenes, determined by mean square error between target image and source image (Fig. \ref{fig:driving_results}c). All non-zero magnitudes of brain transfer $\alpha$ led to faster approximation to convergence values, and we also observe that a set of $\alpha$ values ($\alpha = [0.06,0.08,0.1,0.12,0.14,0.16,0.18]$) led to significantly better performance at the end of training (mean loss $ = [0.074,0.074,0.071,0.073,0.072,0.073,0.072]$, respectively). This is compared to the absence of brain transfer, at mean loss $= 0.080$ ($p<0.01$ for $\alpha \in \{0.06,0.12,0.18\}$ and $p<0.001$ for $\alpha \in \{0.08,0.1,0.14,0.16\}$, Wilcoxon rank-sum test, one-sided), (Fig. \ref{fig:driving_results}d). The displayed standard error was computed across 10 initialization seeds. To address the possibility that the network might be learning better with B2M than no-transfer due to noise regularization, we performed 10 control runs for each value of $\alpha$, changing brain embeddings for standard Gaussian noise. We observe that noise-driven learning performs worse than B2M in this task as well (Fig. \ref{fig:driving_results}d).

For visualization purposes, we include examples of original and reconstructed driving scenes with brain transfer, for $\alpha=0.1$ (Fig. \ref{fig:driving_results}e).

\section{Limitations}

The observed improvements in learning performance could still be partially attributable to a regularization effect introduced by structured noise in the neural data, rather than reflecting brain information transfer alone. Future work should explore more refined controlled ablation studies (e.g., using permuted or synthetic neural data with matched noise statistics to disentangle representational transfer from implicit regularization effects).

In a small subset of training runs, Brain Latent Transfer produced instability, occasionally resulting in catastrophic model divergence. This could potentially arise when the neural training data and model training data differ substantially in input distribution, potentially exposing the model to conflicting or out-of-distribution (OOD) latent signals. Addressing this challenge may require more robust alignment techniques, such as OOD detection or reweighting schemes to reconcile mismatched training domains. Additionally, systematic benchmarking across architectures and input domains will be essential for assessing B2M’s scalability.

Furthermore, the benefits of B2M were tested on RNNs and VAEs. It remains unclear whether these findings extend to other architectures, such as reinforcement learning agents, or task modalities beyond vision and memory-based decision making. Further work must be done to establish extended generalization. Systematic benchmarking across architectures and input domains will be essential for assessing B2M’s scalability.

Finally, while B2M improves testing performance, it is not yet determined what portions of information are being transferred and whether the brain-derived embeddings encode high-level abstractions, low-level features, or task-specific biases. Developing tools to interpret and visualize the aligned latent spaces will be useful for understanding the semantic content of the transfer signal.

\section{Ethical considerations}
Despite its promise, this line of work introduces important ethical considerations. Critically, the use of human brain data for training artificial models raises issues of privacy, consent, and data stewardship. All neural data used in this study were anonymized and collected under IRB approval with informed consent, and care must be taken that these standards are upheld even if the economic viability of B2M in large-scale projects is demonstrated. 

Additionally, techniques that align AI systems with neural representations could, in theory, be misused in contexts such as surveillance or cognitive-behavioral manipulation. Although we do not release pretrained models, any future release will include usage terms to prevent misuse in the context of human subjects protection. We encourage the community to proactively discuss ethical governance frameworks and emphasize full transparency, human subjects protection, consent, and user autonomy in downstream applications. Additionally, care must be taken that human participants are compensated fairly for the data they provide for building better models, which could ultimately provide significant economic potential at scale.


\section{Discussion}

In this work, we demonstrate, as a proof of principle, that low-dimensional brain representations can be leveraged as a substrate to improve neural network training in complex cognitive tasks involving sensorimotor processing, memory, and flexible decision-making. Our results contribute to the growing body of literature at the intersection of neuroscience and machine learning, suggesting that brain-derived priors may offer useful guidance for training artificial agents, above and beyond utilizing such agents for explaining variance patterns in neural data.

This approach raises several promising avenues for future exploration. First, future work may systematically investigate the scalability of brain-guided training: what is the sample efficiency of such priors, and how does performance scale with the size, dimensionality, and diversity of the neural embeddings? Understanding these dynamics could enable principled integration of neural data into large-scale machine learning pipelines.

Second, an open question remains regarding the transferability and utility of different brain recording modalities. Given their varying signal-to-noise ratios and spatial/temporal resolution, it is critical to benchmark the relative effectiveness of invasive (e.g., ECoG, depth recordings) versus non-invasive (e.g., fMRI, EEG) data in shaping model representations across cognitive domains.

Third, this work requires a deeper investigation into the geometry of the brain-induced low-dimensional spaces. By characterizing how artificial models restructure their internal representations in response to brain-derived constraints, we can gain insight into both the alignment between biological and artificial computation and the types of inductive biases these embeddings confer.

Fourth, future work can explore modular alignment strategies, in which targeted brain embeddings derived from functionally specific regions (e.g., early visual cortex, hippocampus, prefrontal cortex) are used to selectively shape or pre-train corresponding architectural modules (e.g., CNNs, RNNs, value-based learners). This could provide a biologically grounded path toward more interpretable and modular AI systems, with potential implications for embodied cognition.

More broadly, this work opens a path toward data-efficient brain-aligned training, where sparse but functionally aligned neural recordings guide the acquisition of generalizable representations in artificial agents. By continuing to bridge neuroscience and machine learning through shared principles of representation and computation, we may build novel strategies for systems that learn uniquely human skills.

\section{Author Contributions}
\small
TGA conceptualized and implemented the B2M method. TGA, VL, PS, and NR contributed to the final implementation of the B2M algorithm. TGA and VL deployed B2M for the driving task. VL preprocessed driving task data. TGA deployed B2M for the memory task. HA and SAS conceptualized the memory task. HA, RA, AJW, EB, BYH, and SAS operated recording systems and collected memory task data. HA and RA preprocessed memory task brain and behavioral data.
TGA, VL, BYH, PS, SAS, and NR wrote the manuscript.

\section{Acknowledgements}
The memory task experiment was funded by NIH grant 5U01NS121472-05. We are grateful for the support from the Air Force Office of Scientific Research under award number FA9550-22-1-0337. SAS is a consultant for Boston Scientific, Abbott, Koh Young, Zimmer Biomet, Neuropace; Co-founder of Motif Neurotech.
\normalsize

\newpage

\bibliographystyle{unsrt}
\bibliography{refs} 

\newpage

\section*{Technical Appendix}
\subsection*{Computer Resources}
All experiments were performed on a Lambda Labs server, with the following characteristics: 192 CPUs, AMD Ryzen Threadripper PRO 7995WX 96-Cores, 5390MHz (maximum), graphics card: NVIDIA Corporation AD102GL [RTX 6000 Ada Generation], 512GB RAM, 18 TB HD storage.

On this machine, the 220 memory task runs (RNN: B2M and noise) took 14.16h in total, whereas the 220 driving task runs (VAE: B2M and noise) took 113.59h in total.

\subsection*{Human Subjects}
For the memory-based decision-making task, we collected intracranial data from 6 human epilepsy patients, undergoing seizure monitoring prior to surgery. Patients read the following text on the screen, substituting target-stim1 and target-stim2 for each episode with the actual target stimuli for that episode:

\begin{quote}
    
A scientist is studying different patterns. Your job is to help him. Objects will appear one after another, and he wants you to take a picture when you spot a particular pattern in their sequence. (For example, a Vase followed by a Flower.)"

We'll always tell you what pattern to look for. Every few objects, there will be a new pattern to watch for. When you spot the pattern, press your photograph button to take a picture. Otherwise, press the appropriate button to skip that object. We'll tell you which button is which on each trial. The objects can appear in any order, including several of the same type in a row. A new object will appear every two seconds or so."

"Press 'LEFT' to take a picture, and press 'RIGHT' to skip this object. Wait for the first [target-stim1], then take a picture of it. Then wait for the first [target-stim2], and take a picture of it. Alternate taking pictures of one [target-stim1] and one [target-stim2].
\end{quote}

\begin{figure}[htb!]
  \centering
  \includegraphics[width=0.3\textwidth]{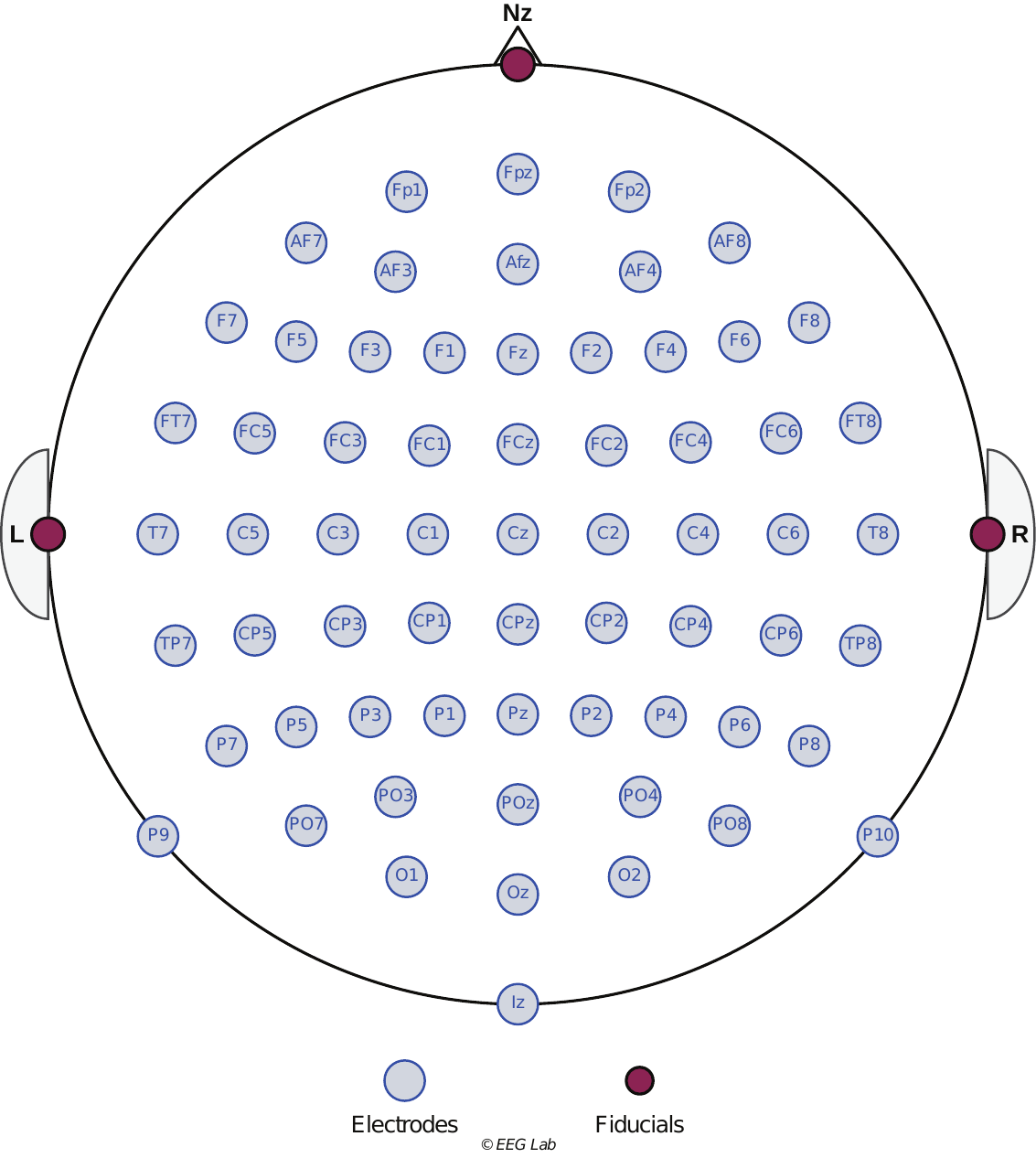}
 \caption{EEG standard contact map. All participants in the VR driving task underwent EEG recording, with electrodes positioned along the same standard contact grid. Nz indicates the direction of the front of the head.}
  \label{fig:eeg_map}
\end{figure}

For the VR driving task, we collected EEG data from 9 healthy human subjects. They had prior driving experience, normal or corrected-to-normal vision, and reported they were not prone to motion sickness. Subjects were verbally instructed to drive a car along a road in VR for a fixed amount of time, avoiding collisions. They were told that collisions would deduct an amount of money from their total reward bonus, which was displayed on the car's dashboard. They could accelerate, brake, and steer with realistic input controls. Subjects were compensated at a rate of $ \$20$/h for 3 hours. EEG contact positioning was the same for all subjects, as displayed in Fig. \ref{fig:eeg_map}

\end{document}